\documentclass{article}
\usepackage[numbers]{natbib}

\usepackage[final]{nips_2017}

\usepackage[utf8]{inputenc} 
\usepackage[T1]{fontenc}    
\usepackage{hyperref}       
\usepackage{url}            
\usepackage{booktabs}       
\usepackage{amsfonts}       
\usepackage{microtype}      
\usepackage{graphicx}
\usepackage{multirow}

\title{Transfer Learning with Binary Neural Networks}

\author{
  Sam Leroux \textbullet \hspace{2pt} Steven Bohez \textbullet \hspace{2pt} Tim Verbelen \textbullet \hspace{2pt} Bert Vankeirsbilck \\ \textbf{Pieter Simoens \hspace{1pt} \textbullet \hspace{2pt}  Bart Dhoedt}\\
  Ghent University - imec, IDLab, Department of Information Technology\\
  Ghent University\\
  Ghent, Belgium \\
  \texttt{sam.leroux@ugent.be} \\
}

\begin{document}

\maketitle

\begin{abstract}
Previous work has shown that it is possible to train deep neural networks with low precision weights and activations. In the extreme case it is even possible to constrain the network to binary values. The costly floating point multiplications are then reduced to fast logical operations. High end smart phones such as Google's Pixel 2  and Apple's iPhone X are already equipped with specialised hardware for image processing and it is very likely that other future consumer hardware will also have dedicated accelerators for deep neural networks. Binary neural networks are attractive in this case because the logical operations are very fast and efficient when implemented in hardware.
We propose a transfer learning based architecture where we first train a binary network on Imagenet and then retrain part of the network for different tasks while keeping most of the network fixed. The fixed binary part could be implemented in a hardware accelerator while the last layers of the network are evaluated in software. We show that a single binary neural network trained on the Imagenet dataset can indeed be used as a feature extractor for other datasets.
\end{abstract}

\section{Introduction}
Deep learning really took off in 2012 when Krizhevsky et al. showed record breaking results on the Imagenet dataset \cite{krizhevsky2012imagenet}. They demonstrated that deep convolutional neural networks trained end to end on large labelled datasets can beat most other techniques for image recognition. Deep learning quickly became the default algorithm for image classification and now even achieves super-human level performance \cite{DBLP:journals/corr/HeZR015}. Deep learning has also revolutionized other fields like speech recognition and natural language processing \cite{lecun2015deep}.
\\
\newline
The two key ingredients needed to successfully apply deep neural networks are large amounts of labelled training data and powerful computing systems such as GPUs. Mobile devices including smartphones, Internet-of-Things (IoT) devices or smart home assistants have very limited processing power because of their intrinsic limitations on size and energy consumption. One possible solution is to offload all computations to the cloud but this introduces a latency and potentially even a privacy risk when sensitive data is processed remotely. 
\\
\newline
There is a considerable amount of active research on techniques to reduce the computational cost of deep learning models. One approach is to prune the network by removing redundant weights. The idea of pruning already goes back to the eighties when LeCun et al. used second order derivatives to calculate the impact of each weight on the loss of the network \cite{NIPS1989_250}. Weights with a small impact are then removed from the network. More recently pruning was used on modern deep neural networks. Song et al. proposed a pruning pipeline where first weights with a small magnitude where removed after which the network was fine-tuned to recover the lost accuracy \cite{han2015learning}. Network pruning is especially effective when a network is used for transfer learning. In transfer learning the model is first trained on a large dataset such as Imagenet and is then fine-tuned on a small domain specific dataset.\\ Because the network was pretrained on a general dataset it will contain convolutional kernels that are not useful for the domain specific dataset. Molchanov et al. introduced a criterion based on a first-order Taylor expansion to decide which feature maps to remove and demonstrated impressive results when used together with transfer learning \cite{molchanov2016pruning}.
\\
\newline
Instead of compressing a trained model it is also possible to train efficient models from scratch. The recently introduced MobileNets \cite{howard2017mobilenets} use depthwise separable convolutions to reduce the computational cost. Depthwise separable convolutions factorize a standard convolution into a depthwise convolution and a 1x1 pointwise convolution. The depthwise convolution applies a single convolution to each input channel while the pointwise convolution combines the information in the different channels. Factorizing a traditional convolution into these two convolutions dramatically reduces the computational cost and size of the network with only a minimal reduction in accuracy.
\\
\newline
Most implementations of deep neural networks use 32 bit floating point numbers for weights and activations. Various works have shown that this is not necessary and that it is possible to use 16 bit \cite{micikevicius2017mixed} or 8 bit \cite{vanhoucke2011improving} numbers. In the extreme case it is even possible to use binary weights and activations. Courbariaux et al. successfully trained convolutional neural networks for image recognition with binary weights and activations \cite{courbariaux2016binarized}. This works surprisingly well for small scale datasets such as CIFAR10 but there is still a large drop in accuracy for larger datasets such as Imagenet. Neural networks with binary weights and activations are attractive because they replace the costly floating point multiplications and additions with bitwise XNORs and left and right bit shifts. These operations are very efficient to implement in hardware.
\\
\newline
Another problem with deep learning is the need for large labelled datasets. Training a new model from scratch requires a large amount of training data. A well known technique is to use transfer learning where a model is first trained on a large dataset like Imagenet and afterwards the last layer is removed and retrained using a small amount of new domain specific training data. Transfer learning works because the first layers in the network learn to detect features such as color transitions and basic shapes that are present in images from different domains \cite{yosinski2014transferable}.

\section{A hybrid hardware-software architecture}
\begin{figure}[h]
 \centering
 \includegraphics[scale=0.43]{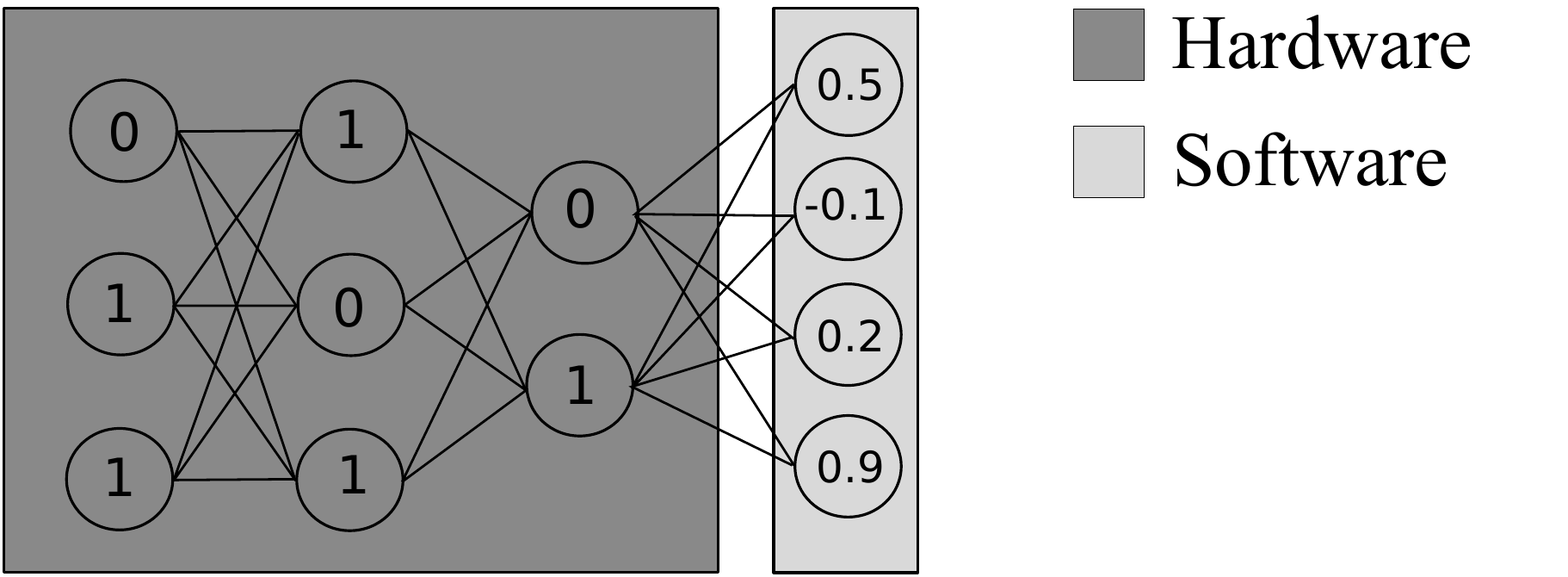}
 \caption{Schematic overview of our proposed architecture. Most layers of a binary neural network trained on Imagenet are embedded in a specialised circuit in hardware. This circuit is used as a fixed feature extractor. The last layer is implemented in software and retrained for different applications.}
 \label{fig:arg}
\end{figure}

We propose a hybrid hardware-software architecture based on this idea of transfer learning. We train a neural network with binary weights and activations on the Imagenet dataset and use this network as a fixed feature extractor that could be optimised on hardware level. The last layer is implemented in software and is evaluated on the CPU (or GPU) of the device. Most of the computations are offloaded to the custom circuit. Since we only need to retrain the last layer of the network it even becomes feasible to train on the device itself instead of offloading the training to the cloud. This is very attractive from a privacy point of view because the training data never leaves the mobile device. Our architecture also allows app developers to embed custom neural networks into their apps. Right now this is often impossible because a typical neural network quickly requires hundreds of megabytes of storage just for the weights. If the device however is equipped with the fixed feature extractor we only need to ship the last (domain specific) layer of the network with the app. A schematic overview of our architecture is shown in Figure \ref{fig:arg}.

\section{Experiments}
In this section we show that it is possible to use a binary neural network trained on Imagenet as a feature extractor for other datasets. We trained a binary version of the Alexnet architecture \cite{krizhevsky2012imagenet} on the ILSVRC2012 training set following the binarization technique of Courbariaux et al. \cite{courbariaux2016binarized}. Our binary network obtains a top 5 accuracy of 67\% (42\% top 1) while the floating point Alexnet obtains a top 5 accuracy of 80\% (57\% top 1). We then retrained the last layer on the domain specific datasets without changing the weights of the other layers. We report results for the three fine-grained image datasets summarized in Table \ref{tbl:datasets}. For all our models we resized the input images such that the longest side was of length 256. During training we took random 224 by 224 pixel crops. For the test set we used center crops. The floating point networks where trained with stochastic gradient descent with momentum. For the binary networks we found that Adam \cite{kingma2014adam} gives better results which is consistent with Courbariaux et al. \cite{courbariaux2016binarized}.

\begin{table}[h]
  \caption{The different datasets used in our experiments. We pretrained our network on the Imagenet dataset and then retrained the last layer on the three smaller domain specific datasets. }
  \label{tbl:datasets}
  \centering
  \begin{tabular}{llll}
    \toprule
    Dataset & Number of classes & Training images & Testing images \\
    \midrule
    ILSVRC2012 & 1000 & 1,200,000 & 50,000\\
    \midrule
    Flowers \cite{Nilsback08} & 102 & 6,149 & 1,020\\
    UCSD Birds \cite{WelinderEtal2010} & 200 & 5,994 & 5,794\\
    MIT Indoor scenes \cite{quattoni2009recognizing} & 67 & 5.360 & 1,340\\
    \bottomrule
  \end{tabular}
\end{table}

The results are summarized in Table \ref{sample-table}. The first part of the table shows the accuracies for the baseline floating point networks, either trained from scratch on the domain specific datasets or fine-tuned from the Imagenet model.
As expected the fine-tuned models consistently outperform the models trained from scratch. The second part of the table shows the results for our binary models. The fine-tuned models again outperform the models trained from scratch which shows that transfer learning also works when the network uses binary weights and activations. For the last layer we can use binary weights and activations (a) but because this layer is evaluated in software in our proposed architecture we also experimented with floating point weights and activations (b). This consistently increases the accuracy on all datasets and pushes the accuracy closer to the accuracy of the floating point models.

\begin{table}[h]
  \caption{The top 1 and top 5 accuracies for each domain specific dataset. We report results for the binary network trained from scratch and for the fine-tuned networks, both with binary weights in the last layer (a) and floating point weights (b). We compare to a floating point baseline.}
  \label{sample-table}
  \centering
  \begin{tabular}{llllllll}
    \toprule
        & & \multicolumn{2}{c}{Flowers}  & \multicolumn{2}{c}{Birds} &  \multicolumn{2}{c}{Indoor scenes}\\
         & & Top 1 & Top 5 & Top 1 & Top 5 & Top 1 & Top 5 \\
    \midrule
    \multirow{2}{*}{FP} & trained from scratch & 71.7\% & 90.5\%  & 32.4\% & 57.2\% & 29.7\% & 56.4\%\\
    & fine-tuned  & 85.4\% & 97.2\% & 50.4\% & 79.6\% & 57.4\% & 86.8\%\\
    \midrule
    \multirow{3}{*}{Binary}& trained from scratch & 63.2\% & 85.1\% & 13.8\% & 32.4\% & 29.0\% & 54.9\%  \\
    & (a) fine-tuned binary last layer & 80.6\% & 94.5\% & 33.9\% & 62.9\% & 46.7\% & 77.4\%\\
    & (b) fine-tuned FP last layer & 84.0\% & 95.5\% & 36.7\% & 63.4\% & 48.3\% & 78.6\% \\
    \bottomrule
  \end{tabular}
\end{table}

\section{Conclusion and future work}
We introduced a hybrid hardware-software architecture where a binary neural network trained on Imagenet can be embedded in a dedicated circuit. The last layer is implemented in software and is retrained for each specific task. We showed that transfer learning works very well for binary neural networks and experimented with a hybrid binary-floating point network where only the last layer uses floating point operations. This is a good trade-off between accuracy and computational cost since most of the computations can be offloaded to the fixed hardware accelerator. In future work we will explore other hybrid architectures. We now only considered finetuning the last layer of the network but we can also retrain more than one layer which should give a higher accuracy but will also incur a higher computational cost since a larger part of the network is evaluated in software.

\section*{Acknowledgments}
We gratefully acknowledge the support of NVIDIA Corporation with the donation of GPU hardware used for this research. Steven Bohez is funded by a Ph.D. grant of the Agency for Innovation by
Science and Technology in Flanders (IWT).
\bibliographystyle{unsrt}
\bibliography{nips_2017} 

\end{document}